\documentclass[10pt,twocolumn,letterpaper]{article}

\usepackage{cvpr}
\usepackage{times}
\usepackage{epsfig}
\usepackage{graphicx}
\usepackage{amsmath}
\usepackage{amssymb}

\usepackage{array}
\usepackage{booktabs}
\usepackage{caption}
\usepackage{multirow}
\usepackage{tabularx}
\usepackage{float}
\usepackage{textcomp}

\cvprfinalcopy 

\def\cvprPaperID{13} 

\ifcvprfinal\fi
\begin{document}

\title{Global Context for Convolutional Pose Machines}

\author{Daniil Osokin\\
IOTG Computer Vision (ICV), Intel Russia\\
{\tt\small daniil.osokin@intel.com}
}

\maketitle

\begin{abstract}
   Convolutional Pose Machine is a popular neural network architecture for articulated pose estimation. In this work we explore its empirical receptive field and realize, that it can be enhanced with integration of a global context. To do so U-shaped context module is proposed and compared with the pyramid pooling and atrous spatial pyramid pooling modules, which are often used in semantic segmentation domain.

   The proposed neural network achieves state-of-the-art accuracy with 87.9\% PCKh for single-person pose estimation on the Look Into Person dataset. A smaller version of this network runs more than 160 frames per second while being just 2.9\% less accurate. Generalization of the proposed approach is tested on the MPII benchmark and shown, that it faster than hourglass-based networks, while provides similar accuracy. The code is available at \url{https://github.com/opencv/openvino\_training\_extensions}\footnote{The full path is: \url{https://github.com/opencv/openvino_training_extensions/tree/develop/pytorch_toolkit/human_pose_estimation}.}.
\end{abstract}

\section{Introduction}

The task of human pose estimation procedure is to recognize human skeleton by localizing position of the keypoints: elbows, wrists, hips, ankles and so on. The obtained pose is usually serves as an important cue in different vision task, such as action recognition, human-computer interaction.

\begin{figure}[h]
    \centering
    \includegraphics[width=0.479\textwidth]{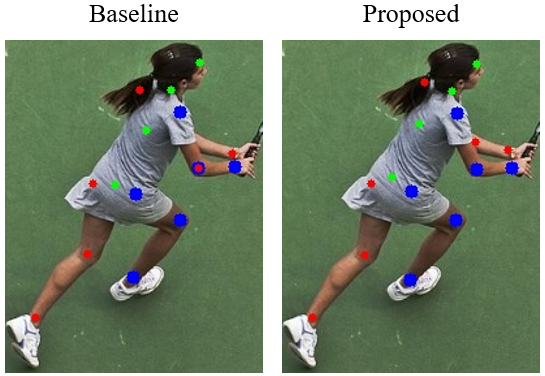}
    \caption{Due to similar appearance the baseline network fails to discriminate between left/right elbow. The proposed network with global context module resolves this misunderstanding. Red dots are the body left-side keypoints, blue dots are the right-side keypoints and green are the central unpaired.}
    \label{fig:ini_samples}
\end{figure}

This paper focuses on the single-person pose estimation, which aims to find the keypoint positions of a single person, that is centered inside a given RGB image patch. This task is hard due to variance in person appearance, body part occlusions, rough estimate of person location and large similarity between left and right keypoint pairs, see Fig. \ref{fig:ini_samples}.
The latter requires capturing full body structure to distinguish between  left and right sides. Moreover, image can contain multiple people, e.g. baseball players or dancers, so the decision about keypoints of which person estimate need to be performed by analyses of the whole scene context.

The noticeable improvement in this domain is connected with usage of convolutional neural networks (CNNs) \cite{tompson2014joint, chen2017adversarial, newell2016stacked, tang2018deeply, toshev2014deppose, wei2016cpm, yang2017pyramid}. The one of pioneering works \cite{toshev2014deppose} proposes the idea of iterative refinement of predictions: pass initial estimate of keypoint coordinates to subsequent estimation stage, which refines initial coordinates. These refined coordinates are passed to the next refinement stage to get more accurate result and so on. The work \cite{tompson2014joint} formulates pose estimation as a regression of spatial heatmaps with peaks centered in keypoint locations. Stacked hourglass network \cite{newell2016stacked} follows both ideas and proposes neural network architecture, which combines features across different scales, by repeated bottom-up and top-down inference, allowing network to maintain spatial resolution and see global image context. A number of works extend this approach by proposing various modifications to the original architecture \cite{zhang2019human, ke2018multi, tang2018deeply, yang2017pyramid}. Another network architecture is convolutional pose machines (CPM), proposed in \cite{wei2016cpm}. It also contains multiple stages performing keypoint heatmaps refinement, but uses several consecutive convolutions with large kernel to increase theoretical receptive field and capture information at global scale. 

In this paper we estimate the {\it empirical} receptive field of CPM-based neural network architecture, and show that it can be enhanced with addition of global context module into network architecture. The proposed network has the complexity 21.3 billions floating-point operations (GFLOPs) and achieves 87.9\% PCKh \cite{andriluka14cvpr} for single-person pose estimation on the Look Into Person (LIP) dataset \cite{gong2017look} and competitive result on MPII Single person pose estimation benchmark \cite{andriluka14cvpr}.

Ours contributions summarized in follows:
\begin{itemize}
\item We evaluate popular global context modules from semantic segmentation domain for pose estimation task and propose U-shaped context module for the receptive field enhancement of CPM-based neural network.
\item We propose an augmentation technique, which mimics body parts occlusion for robust pose estimation.
\item We compare inference time of proposed CPM-based network with hourglass-based networks and analyze bottleneck in performance.
\end{itemize}


\section{Related work}

\subsection{Global context modules in semantic segmentation}

The goal of semantic segmentation is to assign a category label for each pixel in the image. The one of the challenges in this task is caused by the existence of objects at multiple scales. To find objects at coarse scale CNN performs series of feature maps downsampling, however small objects might be missed on feature maps with low resolution.

\begin{figure}[h]
    \centering
    \includegraphics[width=0.45\textwidth]{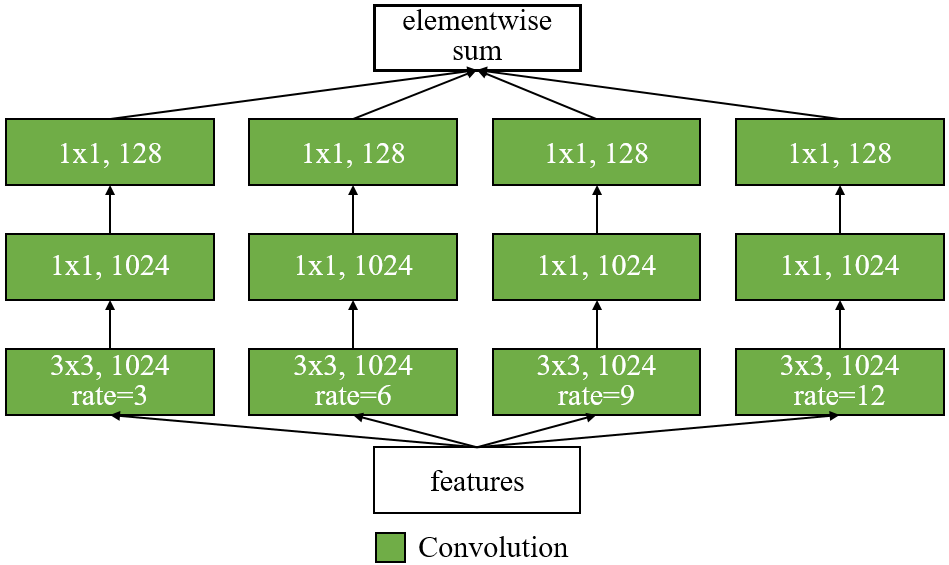}
    \caption{The atrous spatial pyramid pooling module architecture.}
    \label{fig:aspp}
\end{figure}

To tackle this issue \cite{chen2015deeplab} proposed to remove some feature downsamplings and use {\it atrous} convolutions \cite{holschneider2016wavelets, yu2016ms} in subsequent layer to preserve the receptive fileld. Atrous convolution {(or dilated convolution)} can be viewed as:

\[y[i]=\sum_{k}x[i+r\cdot k]\,w[k]\]

\noindent where {\it r} is the sampling rate parameter {(or dilation)}. In other words, this is equivalent to convolving the input $x$ with upsampled filter, obtained by inserting $r-1$ zeros between two consecutive filter values along each spatial dimension. In \cite{chen2018deeplab} was proposed {\it atrous spatial pyramid pooling} {(ASPP)} module, which resamples given feature layer at multiple rates by the atrous convolutions with different value of sampling rate, see Fig. \ref{fig:aspp}. It allows capturing objects and useful image context at multiple scales, because it contains filters, which have complementary field of view.

\begin{figure}[h]
    \centering
    \includegraphics[width=0.45\textwidth]{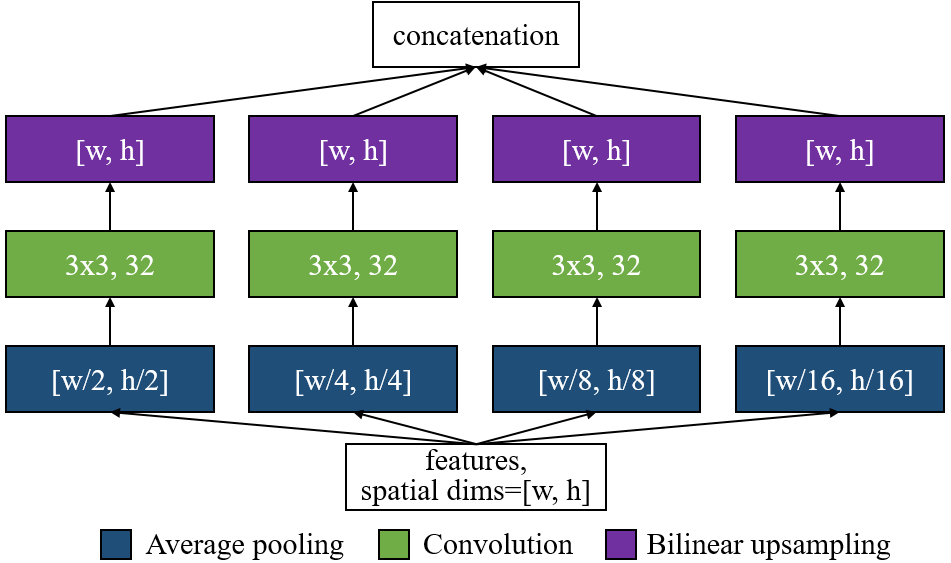}
    \caption{The pyramid pooling module architecture.}
    \label{fig:ppm}
\end{figure}

In the work \cite{zhao2018pspnet} proposed {\it pyramid pooling} module to incorporate global scene context. Instead of dilated convolutions, it utilizes pooling layers with progressively increasing kernel sizes, which captures useful information at different scales, see Fig. \ref{fig:ppm}. After the pooling, resulting feature maps are upsampled to the input resolution and concatenated.

In this paper we evaluate both pyramid pooling module and atrous spatial pyramid pooling module.

\subsection{Single-person human pose estimation}

All modern human pose estimation methods use CNNs as the main building block. The dominant approach for human pose estimation is to estimate keypoint coordinates by performing regression of heatmaps with peaks at keypoints locations. 

Many of the state-of-the-art works base their neural network architecture on original stacked hourglass design \cite{newell2016stacked}. The work \cite{yang2017pyramid} proposed Pyramid Residual Module to enhance the invariance in scales of input features. It learns convolutional filters on various scales of input features in multiple network branches. These features are then aggregated together and fed as an input to the hourglass network. In the work \cite{tang2018deeply} proposed the using of deeply learned compositional models to characterize the complex relationships among body parts. The work \cite{ke2018multi} proposed to train hourglass networks with multi-scale supervision, the keypoint heatmaps from multiple scales are fused to determine the final pose. Instead of the accuracy improvement, the work \cite{zhang2019fast} adressed the cost-effectiveness issue in order to scale the human pose estimation models to large deployments. It adopted knowledge distillation framework \cite{hinton2015distilling} for human pose estimation task and distilled the knowlege from the original neural network with 8-stacked hourglass modules to the neural network with 4-stacked hourglass modules with the reduced channels number.

Our work is based on the CPM approach, which is different to hourglass-based networks in prior works. As shown in \cite{cao2017paf}, this approach allows neural network inference at 15 frames per second (fps) for multi-person pose estimation, thus suitable for real-time  processing.

\section{Baseline network}

\subsection{Design}
\begin{figure*}[t]
    \centering
    \includegraphics[width=0.8\textwidth]{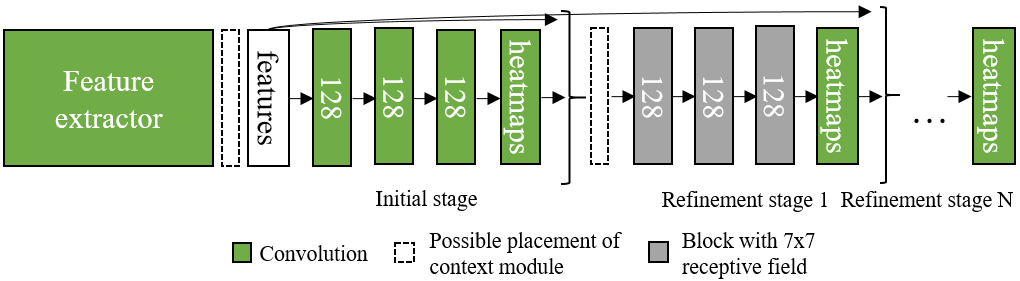}
    \caption{Baseline network architecture. Dotted rectangles are possible positions of context module: right after the feature extractor or in the each refinement stage.}
    \label{fig:net_arch}
\end{figure*}

Our baseline network design is based on the neural network architecture form  \cite{osokin2018lightweight_openpose} and schematically illustrated in Fig.\ \ref{fig:net_arch}. The network starts from the MobileNet \cite{howard2017mobilenets} feature extractor, cut to {\it conv5\_5} layer, and initialized with pre-trained on ImageNet \cite{russakovsky2014imagenet} weights. To maintain spatial resolution of output feature maps a dilated network strategy is applied: stride 2 is removed from the {\it conv4\_2} layer, and dilation parameter of the next convolutional layer is set to 2, so the network regresses the keypoint heatmaps at 8 times downsampled input resolution. After feature extraction, inital keypoints heatmaps are estimated, then 5 refinement stages improve the inital estimation. Each refinement stage consists of the 3 identical blocks, each of them has receptive field of convolutional layer with 7x7 kernel size, but contains less parameters. Сonvlolutional layer with 1x1 kernel size is added to the end of stage to obtain the heatmaps. For the details we refer to the original work.

\subsection{Trainig}

We use PyTorch v1.0 \cite{paszke2017automatic} as a deep learning framework. To train networks Adam optimizer is used with initial learning rate $4\times 10^{-5}$, decreased by a factor of 10 two times on loss platos. For keypoint heatmaps regression the folowing loss is used:

\[L=\frac{1}{N}\sum_{n=1}^{N}\sum_{x, y}(G_n(x, y) - P_n(x, y))^2\]

\noindent where N is a number of keypoint heatmaps and $G_n(x, y), P_n(x, y)$ represent the ground truth and predicted $n$-th heatmap value at $(x, y)$ pixel location. Such loss function is computed for each stage. To measure accuracy PCKh \cite{andriluka14cvpr} metric is used.

\subsubsection{Datasets}
We perform all experiments on the LIP dataset \cite{gong2017look}, which is a popular benchmark for single-person pose estmation. It consists of 50,462 images, which are split into three parts with 30,462 for training, and by 10,000 for validation and testing. Images are obtained by cropping person instances from the Microsoft COCO dataset \cite{lin2014mscoco}. The labels are provided for the 16 keypoint types.

To test the generalization of proposed model, we also train it on MPII dataset \cite{andriluka14cvpr}, which is another popular benchmark for single-person pose estimation. It contains about 25,000 images with over 40,000 people. Annotation also contains 16 different keypoint types.

\subsubsection{Augmentations}
During training, the input image is scaled by the biggest side to 256 pixels and centered in 256x256 pixels image, filled with a fixed color. We perform standard augmentations for pose estimation, specifically scaling image with a factor [0.75, 1.25], rotation in [-40$^\circ$, 40$^\circ$] {([-30$^\circ$, 30$^\circ$] for the MPII dataset)}, horizontal flipping and random permutation of color channels.

\begin{figure}[h]
    \centering
    \includegraphics[width=0.48\textwidth]{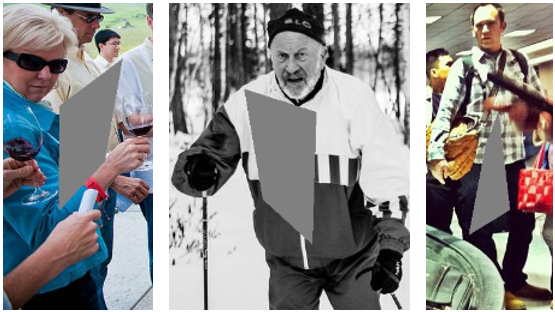}
    \caption{Augmented samples with body masking.}
    \label{fig:aug_samples}
\end{figure}

\noindent {\bf Body masking augmentation.} The LIP dataset contains a lot of persons with occluded body parts, so additional augmentation, which mimics occlusion, may provide more robust model. We propose simple body masking augmentation scheme for such case. To simulate body occlusion, the quadrangle, filled with a uniform color, is drawn at random position near the center of image. The size of quadrangle sides are also random, but not exceeds some pre-defined threshold, we use 30\% of the input image size. Some results of such augmentation are shown in the Fig. \ref{fig:aug_samples}. We also compared proposed occlusion augmentation technique with the one from the work \cite{ke2018multi}, which advises to perform keypoint masking augmentation. Selected set of keypoints is masked with the small image patches form background. Instead of scene background we fill patches with a fixed color. Results are summarized in Table \ref{table:baseline_aug}. 

\begin{table}[h]
\centering
\begin{tabular}{m{11.9em}  c  c}
 \toprule
 Model & PCKh & $\Delta$ \\ 
 \midrule
 Baseline & 82.06 & n/a \\ 
 Baseline + keypoints masking \newline augmentation& 81.96 & -0.1  \\
 Baseline + body masking \newline augmentation (proposed) & {\bf 82.39} & {\bf +0.33} \\
 \bottomrule
\end{tabular}
\caption{Baseline network accuracy on LIP validation subset.}
\label{table:baseline_aug}
\end{table}

\subsection{Empirical receptive field}
\begin{figure}[h]
    \centering
    \includegraphics[width=0.48\textwidth]{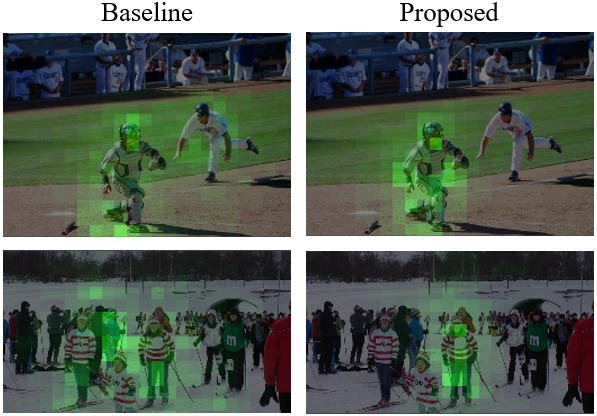}
    \caption{Empirical receptive field for the right hip keypoint. The baseline neural network has difficulty with pose of which person should be estimated. With the addition of global context module the neural network is able to focus on a single person only.}
    \label{fig:multi_erf}
\end{figure}

\begin{figure*}[t]
    \centering
    \includegraphics[width=\textwidth]{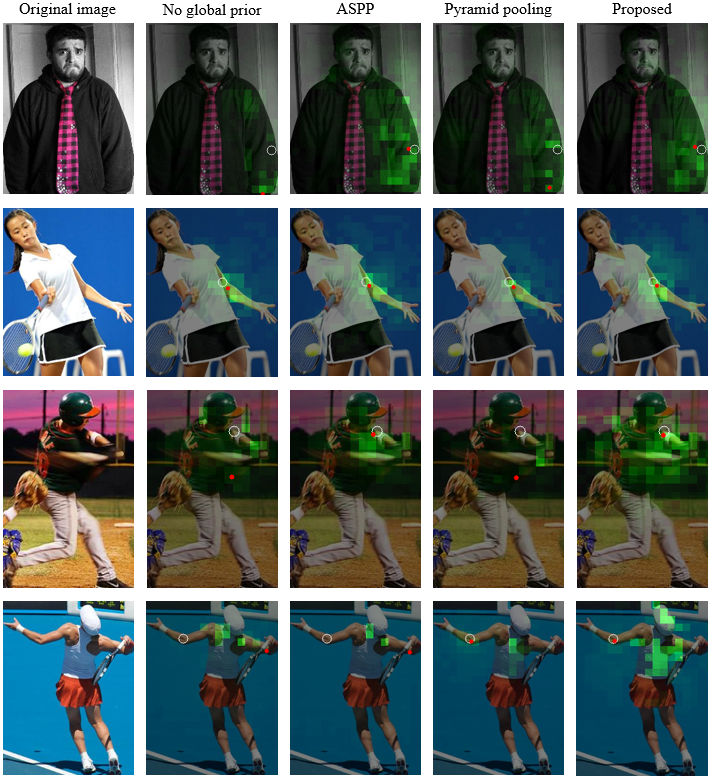}
    \caption{Empirical receptive field and predicted location for the left elbow. It can be seen, that the baseline network without global context focuses only on local region, while with global context module captures left arm or even upper-body to estimate keypoint location. White circle is a ground truth location, red dot is predicted left elbow keypoint for the corresponding context module.}
    \label{fig:l_elb}
\end{figure*}

\begin{figure*}[t]
    \centering
    \includegraphics[width=\textwidth]{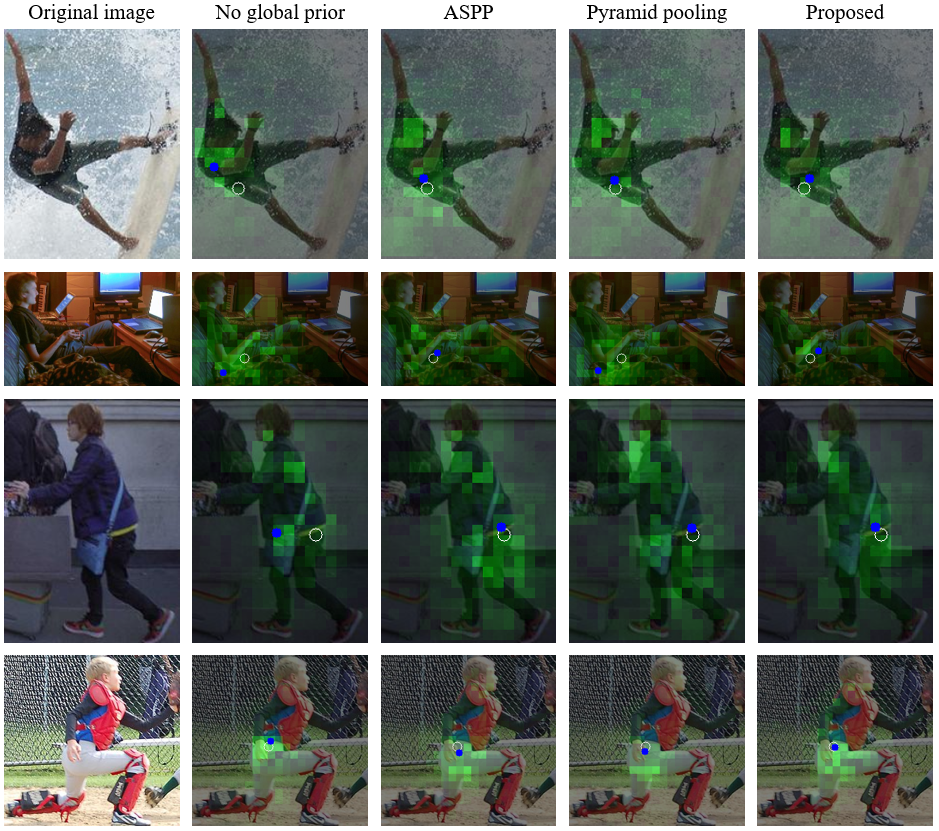}
    \caption{Empirical receptive field and predicted location for the right hip. The baseline network without global context has difficulties with estimation of hip location. Global context modules allow to increase receptive field and use information from almost full pose for location estimation. White circle is a ground truth location, blue dot is predicted right hip keypoint for the corresponding context module.}
    \label{fig:r_hip}
\end{figure*}

Large receptive field is crucial for capturing relationship between pose keypoints \cite{wei2016cpm}. To find the regions of input image, which contributed the most to the keypoint heatmaps, estimation of the {\it empirical} receptive field is performed for layers, which directly regress the heatmaps. Following the work \cite{zhou2015object}, we slide small window of size 11x11 pixels filled with random values over the input image. These images are fed to the neural network and heatmaps are computed. Then they subtracted from the heatmaps, computed from original image, and difference between two heatmaps means that region, masked with random values in original image, is important for heatmap computation. Such regions form the empirical receptive field. Some examples for the baseline network provided in the Fig. \ref{fig:l_elb}, \ref{fig:r_hip}. The green rectangles correspond to empirical receptive field, the saturation is proportional the difference between heatmaps. Visualization shows, that baseline network makes decision based on local region, which does not capture the full pose, and can mislead left-side keypoints with right-side. Another factor, is that it has difficulties in deciding pose of which person should be estimated in case of multiple people inside the image, see Fig.\ \ref{fig:multi_erf}. In the next section we show, that large field of view can tackle this issues.

\section{Receptive field enhancement with global context modules}

To make receptive field larger we propose to incorporate scene information at global scale by adding context modules into the baseline architecture. The one possible way is to add context module to the end of the feature extractor, see Fig. \ref{fig:net_arch}, the left dotted rectangle. However, it was found during experiments, that context module works better if it sees refined heatmaps from previous stage, so it was added to the input of each refinement stage, see Fig.  \ref{fig:net_arch}, the dotted rectangle from the right. We used original ASPP module, with 1024 intermediate channels and [{\it 3, 6, 9, 12}] dilation rates. For the pyramid pooling module pooling kernels with sizes [{\it (w/2, h/2), (w/4, h/4), (w/8, h/8), (w/16, h/16)}] were used, where {\it w} and {\it h} is the width and height of input feature maps correspondingly.
\\
\\
\\

\begin{figure}[h]
    \centering
    \includegraphics[width=0.45\textwidth]{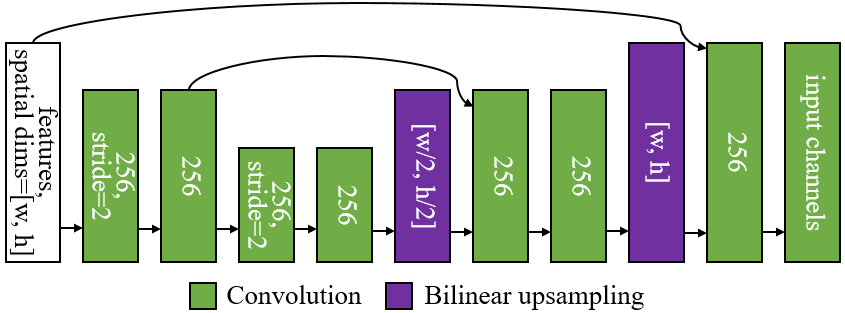}
    \caption{Proposed module for global context embedding.}
    \label{fig:u_shaped}
\end{figure}

Inspired by the succsess of U-shaped {(or encoder-decoder)} neural network architecture \cite{chen2018encoder, iglovikov2018ternaus, ronneberger2018unet}, we propose to use this approach for estimation of global context in human pose estimation task. Specifically, the proposed module consists of multiple convolutions, which downsample input feature maps to the half of its spatial resolution. Feature maps at the lowest resolution are upsampled and concatenated with corresponding features from a higher layer, and so on, until input spatial resolution is reached, see Fig. \ref{fig:u_shaped}.


The results are collected in the Table \ref{table:baseline_context}. It can be seen that the biggest improvement is for hips {(+2.8\%)} keypoints. This is because hips do not have unique appearance features and require understanding of pose keypoints relationships, thus greatly benefit from capturing the full pose by the large field of view.


\begin{table}[h]
\centering
\setlength{\tabcolsep}{2pt}
\begin{tabularx}{0.48\textwidth}{l*{9}{c}}
 \toprule
 Model & Head & Sho. & Elb. & Wri. & Hip & Knee & Ank. & PCKh \\
 \midrule
 Baseline & 93.3 & 89.0 & 84.1 & 80.3 & 70.6 & 77.7 & 78.6 & 82.4  \\
 Baseline & \multirow{2}{*}{93.6} & \multirow{2}{*}{90.1} & \multirow{2}{*}{85.0} & \multirow{2}{*}{81.3} & \multirow{2}{*}{72.6} & \multirow{2}{*}{79.7} & \multirow{2}{*}{79.3} & \multirow{2}{*}{83.5}  \\
 + ASPP\\
 Baseline & \multirow{3}{*}{93.3} & \multirow{3}{*}{90.0} & \multirow{3}{*}{84.9} & \multirow{3}{*}{81.2} & \multirow{3}{*}{71.9} & \multirow{3}{*}{78.9} & \multirow{3}{*}{79.1} & \multirow{3}{*}{83.2}  \\
 + pyramid \\
 \hspace{0.8em}pooling \\
 Baseline & \multirow{2}{*}{\bf 93.7} & \multirow{2}{*}{\bf 90.4} & \multirow{2}{*}{\bf 85.5} & \multirow{2}{*}{\bf 82.1} & \multirow{2}{*}{\bf 73.4} & \multirow{2}{*}{\bf 80.2} & \multirow{2}{*}{\bf 80.2} & \multirow{2}{*}{\bf 84.0}  \\
 + U-shaped\\
\bottomrule
\end{tabularx}
\caption{Evaluation of different global context modules on LIP validation subset.}
\label{table:baseline_context}
\end{table}

Some qualitative results are visualized in Fig.\ \ref{fig:l_elb}, \ref{fig:r_hip}. It can be seen, that the lack of information about the whole scene, which corresponds to small receptive field, causes inaccuracies in keypoints localization and even misleading with left/right body side by the baseline. We have noticed, that network can compensate this problems if it is able to capture context for the entire limb, e.g. left arm for left elbow, or almost the full pose for right hip. Addition of the pyramid pooling or ASPP module to the baseline network improves the result and makes receptive field larger. The proposed U-shaped global context module provides more stable result.


\section{Performance ablation study}
\subsection{Experiments setup}
To estimate the performance/accuracy trade-off we evaluated the effect of proposed modifications on the neural network inference speed. Nvidia graphics processing unit (GPU) GTX 1080 Ti was used as test device, and experiments were performed in PyTorch v1.0 built with CUDA 10.1 and cuDNN 7.5. Batch size was set to 1 to test the real-time inference capability.

\subsection{Complexity of global context modules}
Table \ref{table:comparison_context} shows the complexity of the each context module. It worth nothing, that ASPP module is the heaviest one and the pyramid pooling module is extremely lightweight, yet improves the baseline network result. We've tried to make it wider to achieve better accuracy, e.g. use 128 channels number in convolutional layers instead of 32, however accuracy didn't increase. Thus the only proposed U-shaped module allows to achieve the best results.

\begin{table}[h]
\centering
\begin{tabular}{m{8em}  c  c}
 \toprule
 Module & Parameters$\cdot 10^6$ & GFLOPs \\ 
 \midrule
 ASPP & 9.52 & 9.75 \\ 
 Pyramid pooling & 0.2 & 0.07 \\ 
 U-shaped & 6.44 & 2.53 \\ 
 \bottomrule
\end{tabular}
\caption{Complexity of context modules.}
\label{table:comparison_context}
\end{table}

\subsection{Results on the LIP dataset}
Table \ref{table:perf_lip} shows the result on the LIP test set. The {\it Ours} corresponds to the result of proposed model, obtained with averaged keypoint heatmaps from 3 input scales: [0.75, 1.0, 1.25] and flipped image, neural network complexity and inference time are shown for a single scale testing. {\it Ours 2-stage} corresponds to the result of proposed model with only 2 stages of keypoint heatmaps estimation: initial and one refinement, testing was performed on a single scale of input image.

\begin{table}[h]
\centering
\setlength{\tabcolsep}{2pt}
\begin{tabularx}{0.405\textwidth}{l c c c}
 \toprule
 \multirow{2}{*}{Method} & \multirow{2}{*}{PCKh} & \multirow{2}{*}{Hip}  & Inference \\
 &&&time, ms \\
 \midrule
  Hybrid Pose Machine &77.2 & 69.2 & n/a \\
  BUPTMM-POSE  &80.2 & 68.5 & n/a \\
  Chou et al., CVPR'17\cite{chou2017self} & 87.4 & 75.7 & 33 \\
  Nie et al., CVPR'18\cite{nie2018pil} & 87.5 & 75.9 & 60 \\
  Ours & {\bf 87.9} & {\bf 78.0} & {\bf 18} \\
  Ours 2-stage & 85.0 & {74.1} & {\bf 6} \\
\bottomrule
\end{tabularx}
\caption{Results on the LIP test set.}
\label{table:perf_lip}
\end{table}

The first two methods combine predictions from multiple neural networks. The method \cite{nie2018pil}\footnote{Implementation is taken from: \url{https://github.com/NieXC/pytorch-pil}.} is based on hourglass network and uses additional information from human parsing as a valuable cue for pose estimation. The work \cite{chou2017self} used 4-stacked hourglass network and adversarial learning paradigm to achieve plausible human body configurations. These methods are complementary to the proposed one, and may further boost the accuracy. For hips, the hardest type of keypoints to locate, our method outperforms the hourglass-based competitors on more than 2\%, thus proves the efficiency of proposed U-shaped module in capturing of full pose context.

\subsection{Inference on CPU}
To further validate real-time inference capability, we test our 2-stage version on the Intel\textregistered\ Core\textsuperscript{TM} i7-6850K CPU with Intel\textregistered\ OpenVINO\textsuperscript{TM} Toolkit as inference framework. The network runs at 19.5 frames per second on the CPU, thus may be suitable for real-time processing even on embedded devices without a discrete GPU.

\subsection{Comparison with hourglass-based networks on the MPII dataset}
We have compared proposed CPM-based neural network against hourglass-based neural networks on the MPII dataset. For better results the size of heatmaps was set to 64 in our network. The results\footnote{Since works \cite{ke2018multi, yang2017pyramid} can be reduced to the original 8-stacked hourglass neural network, their complexity and inference time cannot be faster than it, so for them we use the original hourglass network measurements with a '+' symbol} are shown in the Table \ref{table:perf_comparison}. It can be seen, that the proposed network improved the result of original work \cite{wei2016cpm} on 3\%, while being $\sim$25\% faster. Furthermore, it is faster than hourglass-based methods and has similar accuracy.

To compare the real-time inference capability of hourglass-based methods, we profiled the fastest hourglass-based network \cite{zhang2019fast} and our with Nvidia profiler. The profile showed, that despite higher theoretical complexity, our network is faster. The bottleneck in hourglass-based networks is convolutional layers with kernel size 1x1, which take the most of execution time. Such multiple consecutive convolutions {(which form residual block \cite{he2016resnet} repeated many times in hourglass module)} just cannot effectively load GPU. However they make neural network deeper, thus forcing to do more memory-bound read/write operations, which consume the time. 

\begin{table}[h]
\centering
\setlength{\tabcolsep}{2pt}
\begin{tabularx}{0.42\textwidth}{l c c c}
 \toprule
 \multirow{2}{*}{Method} & \multirow{2}{*}{AUC} & Inference & \multirow{2}{*}{PCKh} \\
 &&time, ms \\
 \midrule
  Newell et al., ECCV'16\cite{newell2016stacked} & 62.9 & 64 & 90.9 \\
  Yang et al., ICCV'17\cite{yang2017pyramid} & \textbf{64.2} & 64+ & 92.0 \\
  Ke et al., ECCV'18\cite{ke2018multi} & 63.8 & 64+ & {\bf 92.1} \\
  Zhang et al., CVPR'19\cite{zhang2019fast} & 63.5 & 33 & 91.1 \\
  \midrule
  Wei et al., CVPR'16\cite{wei2016cpm} & \multirow{2}{*}{n/a} & \multirow{2}{*}{31} & \multirow{2}{*}{87.9} \\
  (original CPM) \\
  Ours & \textbf{64.2} & {\bf 24.5} & 90.9 \\
\bottomrule
\end{tabularx}
\caption{Comparison of the proposed CPM-based neural network with hourglass-based networks on MPII test set.}
\label{table:perf_comparison}
\end{table}

\section{Conclusion}

We have studied the receptive field of CPM-based neural network for the single-person pose estimation task. It was shown, that empirical receptive field  can be enhanced with the proposed U-shaped context module, which adds the information about global scene context to the feature maps. This allowed network better differentiate between left and right types of keypoints and reduce ambiguities about which pose to estimate in case of multiple persons inside the image.

We have proposed augmentation technique, that simulates body occlusion by masking, which improved prediction results for the occluded keypoints.

We have analyzed real-time inference capability of the proposed network and compare it to widely-used hourglass-based network architecture. It was found, that convolutional layers with kernel size 1x1 take the most of execution time for hourglass-based networks. Such layers cannot effectively utilize GPU and make network inference slower.

Our network achives 87.9\% PCKh for single-person pose estimation on the Look Into Person dataset and runs 55 fps on the Nvidia 1080 Ti GPU. 
The 2-stage version of the network runs 160+ fps, thus we believe it can be suitable for the real-world applications.

{\small
\bibliographystyle{ieee_fullname}
\bibliography{gccpm}

\begin{thebibliography}{10}\itemsep=-1pt

\bibitem{andriluka14cvpr}
M. Andriluka, L. Pishchulin, P. Gehler, and B. Schiele.

\bibitem{cao2017paf}
Z. Cao, T. Simon, S.-E. Wei, and Y. Sheikh.
\newblock Realtime multi-person 2d pose estimation using part affinity fields.
\newblock In {\em CVPR}, 2017.

\bibitem{chou2017self}
C.-J. Chou, J.-T. Chien, and H.-T. Chen.
\newblock Self adversarial training for human pose estimation.
\newblock In {\em CVPR-W}, 2017.

\bibitem{howard2017mobilenets}
A.~G.~Howard et al.
\newblock Mobilenets: efficient convolutional neural networks for mobile vision
  applications.
\newblock {\em arXiv preprint arXiv:1704.04861}, 2017.

\bibitem{paszke2017automatic}
A.~Paszke et al.
\newblock Automatic differentiation in pytorch.
\newblock In {\em NIPS-W}, 2017.

\bibitem{zhou2015object}
B.~Zhou et al.
\newblock Object detectors emerge in deep scene cnns.
\newblock In {\em ICLR}, 2015.

\bibitem{zhao2018pspnet}
H.~Zhao et al.
\newblock Pyramid scene parsing network.
\newblock In {\em CVPR}, 2017.

\bibitem{zhang2019human}
H.~Zhang et al.
\newblock Human pose estimation with spatial contextual information.
\newblock {\em arXiv preprint arXiv:1503.02531}, 2019.

\bibitem{tompson2014joint}
J.~Tompson et al.
\newblock Joint training of a convolutional network and a graphical model for
  human pose estimation.
\newblock In {\em NIPS}, 2014.

\bibitem{gong2017look}
K.~Gong et al.
\newblock Look into person: self-supervised structure-sensitive learning and a
  new benchmark for human parsing.
\newblock In {\em CVPR}, 2017.

\bibitem{chen2015deeplab}
L.-C.~Chen et al.
\newblock Semantic image segmentation with deep convolutional nets and fully
  connected crfs.
\newblock In {\em ICLR}, 2015.

\bibitem{chen2018deeplab}
L.-C.~Chen et al.
\newblock Deeplab: semantic image segmentation with deep convolutional nets,
  atrous convolution, and fully connected crfs.
\newblock In {\em TPAMI}, 2018.

\bibitem{chen2018encoder}
L.-C.~Chen et al.
\newblock Encoder-decoder with atrous separable convolution for semantic image
  segmentation.
\newblock In {\em ECCV}, 2018.

\bibitem{holschneider2016wavelets}
M.~Holschneider et al.
\newblock A real-time algorithm for signal analysis with the help of the
  wavelet transform.
\newblock {\em Wavelets: timeFrequency Methods and Phase Space}, 1989.

\bibitem{russakovsky2014imagenet}
O.~Russakovsky et al.
\newblock Imagenet large scale visual recognition challenge.
\newblock 2014.

\bibitem{lin2014mscoco}
T.-Y.~Lin et al.
\newblock Microsoft coco: common objects in context.
\newblock In {\em ECCV}, 2016.

\bibitem{iglovikov2018ternaus}
V.~Iglovikov et al.
\newblock Ternausnetv2: fully convolutional network for instance segmentation.
\newblock In {\em CVPR-W}, 2018.

\bibitem{nie2018pil}
X.~Nie et al.
\newblock Human pose estimation with parsing induced learner.
\newblock In {\em CVPR}, 2018.

\bibitem{chen2017adversarial}
Y.~Chen et al.
\newblock Adversarial posenet: a structure-aware convolutional network for
  human pose estimation.
\newblock In {\em ICCV}, 2017.

\bibitem{he2016resnet}
K. He, X. Zhang, S. Ren, and J. Sun.
\newblock Deep residual learning for image recognition.
\newblock In {\em CVPR}, 2016.

\bibitem{hinton2015distilling}
G. Hinton, O. Vinyals, and J. Dean.
\newblock Distilling the knowledge in a neural network.
\newblock {\em arXiv preprint arXiv:1503.02531}, 2015.

\bibitem{ke2018multi}
H.~Qi S.~Lyu L.~Ke, M.-C.~Chang.
\newblock Multi-scale structure-aware network for human pose estimation.
\newblock In {\em ECCV}, 2018.

\bibitem{newell2016stacked}
A. Newell, K. Yang, and J. Deng.
\newblock Stacked hourglass networks for human pose estimation.
\newblock In {\em ECCV}, 2016.

\bibitem{osokin2018lightweight_openpose}
D. Osokin.
\newblock Real-time 2d multi-person pose estimation on cpu: lightweight
  openpose.
\newblock 2018.

\bibitem{ronneberger2018unet}
O. Ronneberger, P. Fischer, and T. Brox.
\newblock U-net: Convolutional networks for biomedical image segmentation.
\newblock In {\em MICCAI}, 2015.

\bibitem{tang2018deeply}
W. Tang, P. Yu, and Y. Wu.
\newblock Deeply learned compositional models for human pose estimation.
\newblock In {\em ECCV}, 2018.

\bibitem{toshev2014deppose}
A. Toshev and C. Szegedy.
\newblock Deeppose: Human pose estimation via deep neural networks.
\newblock In {\em CVPR}, 2014.

\bibitem{wei2016cpm}
S.-E. Wei, V. Ramakrishna, T. Kanade, and Y. Sheikh.
\newblock Convolutional pose machines.
\newblock In {\em CVPR}, 2016.

\bibitem{yang2017pyramid}
W. Yang, S. Li, W. Ouyang, H. Li, and X. Wang.
\newblock Learning feature pyramids for human pose estimation.
\newblock In {\em ICCV}, 2017.

\bibitem{yu2016ms}
F. Yu and V. Koltun.
\newblock Multi-scale context aggregation by dilated convolutions.
\newblock In {\em ICLR}, 2016.

\bibitem{zhang2019fast}
F. Zhang, X. Zhu, and M. Ye.
\newblock Fast human pose estimation.
\newblock In {\em CVPR}, 2019.

\end{thebibliography}
}

\end{document}